# Analyse du besoin en assistance robotique dans la chirurgie de l'oreille

G. Michel[a,b], D. Chablat[b], P. Bordure[a], F. Péchereau[c], P. Schegg[c]

a. CHU de Nantes, 5 Allée de l'Île Gloriette, 44093 Nantes
{guillaume.michel@chu-nantes.fr, philippe.bordure@chu-nantes.fr}
b. Laboratoire des Sciences du Numérique de Nantes (LS2N), UMR CNRS 6004,
1 rue de la Noë, 44321 Nantes, France, damien.chablat@cnrs.fr
c. Ecole Centrale de Nantes,
1 rue de la Noë, 44321 Nantes
{felix.00729@gmail.com, pierre.schegg@gmail.com}

## Résumé :

*La chirurgie de l'oreille présente quelques spécificités par rapport à d'autres spécialités chirurgicales. L'espace anatomique est un espace de petite taille, avec de nombreuses structures anatomiques à respecter, de natures diverses comme les osselets ou le nerf facial. L'abord de cet espace anatomique nécessite l'utilisation de microscope ou d'endoscope. Le microscope permet de garder les deux mains du chirurgien libre mais ne donne qu'une vision directe. L'endoscope contraint le chirurgien à n'avoir qu'une main pour ses instruments, mais donne une vision « fish-eye ». L'essor de l'endoscopie ces dernières années est tel que de nombreux dispositifs d'assistance au chirurgien apparaissent sur le marché : le Robotol, initialement premier robot chirurgical otologique destiné à effectuer certains mouvements avec précision, peut tenir un endoscope. L'Endofix Exo est un bras non robotisé maintenant un endoscope. Ces deux dispositifs nécessitent la main du chirurgien pour les mobiliser. Aucun dispositif robotisé ne permet à l'heure actuelle de maintenir l'endoscope et de le diriger de façon autonome tandis que le chirurgien conserve ses deux mains libres pour travailler, tout comme lorsqu'il travaille sous microscope. L'objectif de notre travail est de définir les besoins spécifiques à l'assistance en chirurgie otologique.*

## Abstract:

*Otologic surgery has some specificities compared to others surgeries. The anatomic working space is small, with various anatomical structures to preserve, like ossicles or facial nerve. This requires the use of microscope or endoscope. The microscope let the surgeon use both hands, but allows only direct vision. The endoscope leaves only one hand to the surgeon to use his tools, but provides a "fish-eye" vision. The rise of endoscopy these past few years has led to the development of numerous devices for the surgeon: the Robotol, first otological robot designed to performed some movements and hold an endoscope, or the Endofix Exo. Both devices need the hand of the surgeon to be moved. No robotic device allows the endoscope to be directed autonomously while the surgeon keeps both hands free to work, just like when he is working with a microscope. The objective of our work is to define the specific needs of the otological assistance surgery.*

**Mots clefs : robotique ; chirurgie ; otologie ; endoscopie.**

# 1 Introduction

Dans le domaine de la chirurgie de l'oreille, et plus largement de la microchirurgie, plusieurs difficultés sont rencontrées par le chirurgien. L'oreille moyenne est une entité anatomique de faible volume avec de multiples éléments fragiles à ne pas léser. Les opérations sont traditionnellement réalisées sous loupes binoculaires ce qui permet au chirurgien d'utiliser ses deux mains, pour un micro-instrument et un outil d'aspiration. De manière plus récente, le développement de la chirurgie otologique par endoscopie permet une meilleure vision de zones difficiles d'accès [1,2]. Cependant, le maintien de l'endoscope par le chirurgien ne lui permet d'opérer qu'avec un instrument à la fois. Il existe donc une attente forte de la part des chirurgiens otologiques pour un système de co-activité afin de l'assister dans son geste chirurgical. Actuellement, plusieurs systèmes robotisés sont développés, témoignant de l'engouement pour l'assistance robotique en micro-chirurgie. L'équipe du Pr Sterkers [3] a développé un robot télé-opéré à six degrés de libertés pouvant être utilisé avec le microscope opératoire. Le design n'était pas initialement conçu comme une assistance à la chirurgie endoscopique, et sa manipulation nécessite l'utilisation d'une main. D'autres systèmes permettent de réaliser une partie précise de la chirurgie, programmée par le chirurgien et effectuée par le robot : insertion d'une électrode dans la cochlée [4], fraisage d'une mastoïdectomie [5], etc …

La problématique est donc d'améliorer la sécurité du geste dans cet environnement à risque, en assistant le chirurgien principalement dans le cadre de l'utilisation de l'endoscopie. L'objectif de ce travail est de concevoir un système robotisé permettant d'assister le chirurgien à l'aide d'une troisième main, maintenant l'endoscope et suivant les gestes du chirurgien. L'objectif du travail présenté dans cet article est d'introduire l'espace de travail du robot utilisé ainsi que les modalités d'usage de ce robot.

# 2 Caractérisation de l'espace de travail

## 1 Définition de l'espace de travail

L'espace de travail est constitué de la caisse de l'oreille moyenne et du conduit auditif externe. L'endoscopie est le plus souvent utilisée pour le traitement de pathologies affectant ces deux zones anatomiques. Il est possible d'utiliser l'endoscope au niveau de la mastoïde, jusqu'au conduit auditif interne, mais cet espace est plus vaste, variable en fonction du fraisage réalisé par le chirurgien et donc moins contraignant a priori. Cet espace de travail a plusieurs particularités : d'une part, il est de taille variable selon les sujets en physiologie. Il n'y aurait pas de différence significative selon le sexe, l'âge ou le côté de l'oreille [6]. D'autre part, en pathologie, il peut varier d'une absence complète de ces zones (aplasie) à un volume étendu à volonté par le chirurgien (en carcinologie par exemple).

Dans la littérature, les analyses sont le plus souvent radiologiques et portent sur le conduit auditif externe, les osselets ou la mastoïde. Ainsi, l'équipe parisienne développant le Robotol [3] s'est basé sur 12 scanners pour mesurer un cylindre moyen correspondant au conduit auditif externe et à la partie visible de l'oreille moyenne.

Dillon [5], sur modèles cadavériques, et Cros [7], à partir de 10 scanners, se sont intéressés à la mastoïde uniquement ; mais la mastoïde n'est pas la zone de travail privilégiée pour la chirurgie sous endoscopie, et peut être élargie à la demande par fraisage. Pacholke [8] retrouve un volume moyen de l'oreille moyenne de 0.58 cm$^3$ à partir de 15 scanners, avec une dimension axiale maximum de 1.57 cm, alors que Mas [9] l'évalue entre 5.25 et 6.22 cm$^3$ à partir de 18 scanners. La plus grande étude retrouvée à partir de 100 scanners [6] évalue le volume du conduit auditif externe à 1.4 mL et celui de l'oreille moyenne à 1.1 mL. Ce volume diminue significativement lors de présence d'une otite chronique. Au total, les données sont très variables selon les études, et il n'existe pas de mesure géométrique de l'oreille moyenne. Il nous est donc apparu important de réaliser un atlas géométrique afin de mieux définir notre espace de travail. Cette étude est basée sur des scanners des rochers d'une population d'âge variable (n=16, patients de 2 à 79 ans). Les mesures ont porté sur les trois axes de la caisse du tympan, de l'hypotympanum à l'attique, mais aussi du conduit auditif externe, à la jonction conduit osseux-conduit fibro-cartilagineux et au niveau du sulcus. Les mesures moyennes ainsi que les valeurs extrêmes sont présentées en **Figure 1**. La totalité des données est présentée dans le **tableau 1.**

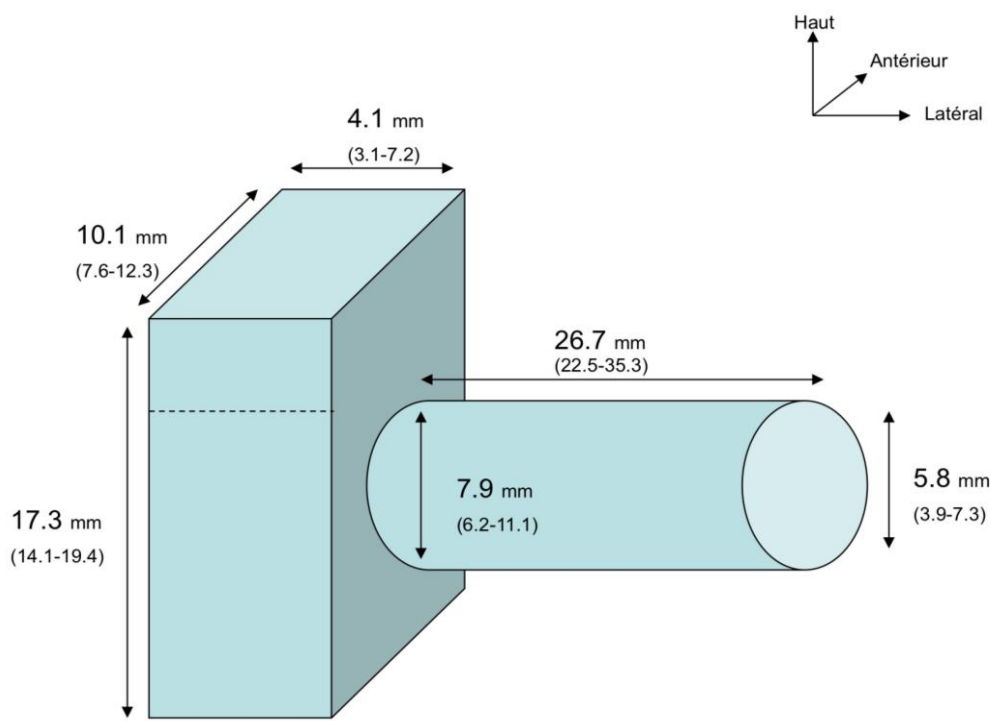

***Figure 1*** : *Tailles moyennes de la caisse du tympan et du conduit auditif externe (n=16).*

## 2  Reconstruction tridimensionnelle de l'espace de travail

La reconstruction tridimensionnelle de l'espace de travail a été réalisée d'après scanner, à l'aide du logiciel Carestream Vue PACS v 11.3. Un rendu volumique est possible avec ce logiciel, en réalisant ensuite un plan de coupe sagittal permettant de visualiser à différents niveaux l'espace de travail considéré dans cette étude. Les images présentées dans ce rapport sont des captures exportées en format JPEG via ce logiciel ; mais il reste possible de naviguer dans cette reconstruction volumique afin de se représenter l'espace de travail selon divers angles de vue et plans de coupe. La première image présentée en **Figure 2** représente la caisse du tympan au niveau de sa paroi médiale ; on peut visualiser la taille moyenne de cette cavité mais également ses différents reliefs, variables, qui la composent. La seconde image, plus latérale, permet de visualiser les osselets tels qu'on les découvre en endoscopie ; en microscopie, la partie supérieure (corps de l'enclume et tête du marteau) ne serait pas visible, sauf atticotomie, c'est-à-dire fraisage du mur osseux nommé mur de la logette, que l'on aperçoit au niveau de la troisième image. La quatrième image représente le conduit auditif externe au niveau de sa portion osseuse ; il permet de comprendre l'étroitesse de cette voie d'accès à la caisse du tympan. En effet, les différents outils présentés au chapitre suivant doivent nécessairement passer par ce conduit osseux. Ce conduit peut éventuellement être élargi à l'aide d'un fraisage, mais cela induit un temps chirurgical plus long et une iatrogénie plus grande.

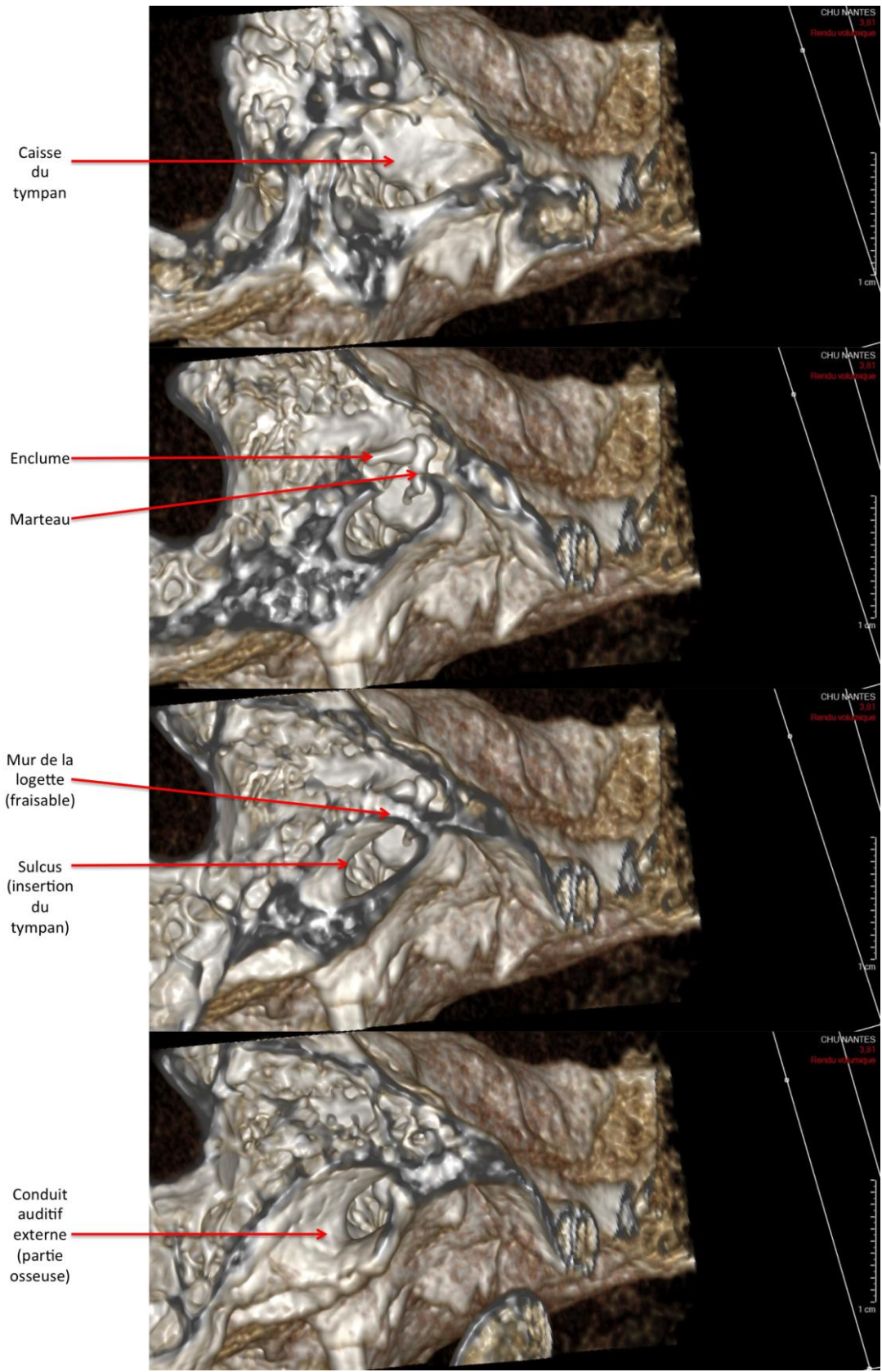

**Figure 2** : Reconstruction 3D d'un rocher droit à partir de scanner : visualisation de l'espace de travail selon l'axe de l'endoscope, en s'éloignant progressivement de la caisse du tympan.

| Age | Diamètre CAE extrémité latérale | Diamètre CAE au sulcus | Longueur CAE | Hauteur OM | Largeur OM | Longueur antéro-postérieure OM |
|---|---|---|---|---|---|---|
| 50 | 7,2 | 9,1 | 26,9 | 19,4 | 11,7 | 5,1 |
| 79 | 5,3 | 11,1 | 34,4 | 15,3 | 12,4 | 6,5 |
| 2 | 4,0 | 9,2 | 22,5 | 14,8 | 11,4 | 6,2 |
| 58 | 3,9 | 6,3 | 28,5 | 15,3 | 8,8 | 6,8 |
| 59 | 7,1 | 9,0 | 35,3 | 14,5 | 12,1 | 4,4 |
| 37 | 6,8 | 8,4 | 25,3 | 18,5 | 11,2 | 5,5 |
| 71 | 6,5 | 10,2 | 27,4 | 15,5 | 11,2 | 5,2 |
| 4 | 4,8 | 8,8 | 23,2 | 15,1 | 12,3 | 6,8 |
| 51 | 4,3 | 6,5 | 26,9 | 14,1 | 9,1 | 7,2 |
| 44 | 7,3 | 9,3 | 31,3 | 19,1 | 10,6 | 4,9 |
| 29 | 6,6 | 6,3 | 28,1 | 15,3 | 9,2 | 4,7 |
| 32 | 6,8 | 6,2 | 23,1 | 14,5 | 7,6 | 4,9 |
| 18 | 5,6 | 6,6 | 25,2 | 14,5 | 10,8 | 6,8 |
| 5 | 5,5 | 7,5 | 27,2 | 16,2 | 11,6 | 6,2 |
| 8 | 5,1 | 6,9 | 22,2 | 16,6 | 8,4 | 4,1 |
| 78 | 5,7 | 6,6 | 26,6 | 15,3 | 8,5 | 3,1 |
| **Moyenne** | **5,79** | **7,85** | **26,75** | **17,35** | **10,10** | **4,10** |
| **Ecart-type** | **1,15** | **1,58** | **3,87** | **1,68** | **1,57** | **1,17** |

*Tableau 1* : *Données issues de l'analyse scannographique concernant 16 patients (CAE : Conduit Auditif Externe ; OM : Oreille Moyenne).*

## 3   Analyse fonctionnel du porte endoscope

A partir de la connaissance de l'environnement dans lequel le système robotisé doit travailler, il est nécessaire de réaliser une analyse fonctionnelle du système dans son environnement. La Figure 3 présente la « bête à cornes » de l'analyse du système. Deux questionnements sont alors possibles :

➔ Comment fixer le porte endoscope

➔ Comment piloter le porte endoscope

C'est sur ces deux questions que nous allons essayer de répondre dans la suite de l'article.

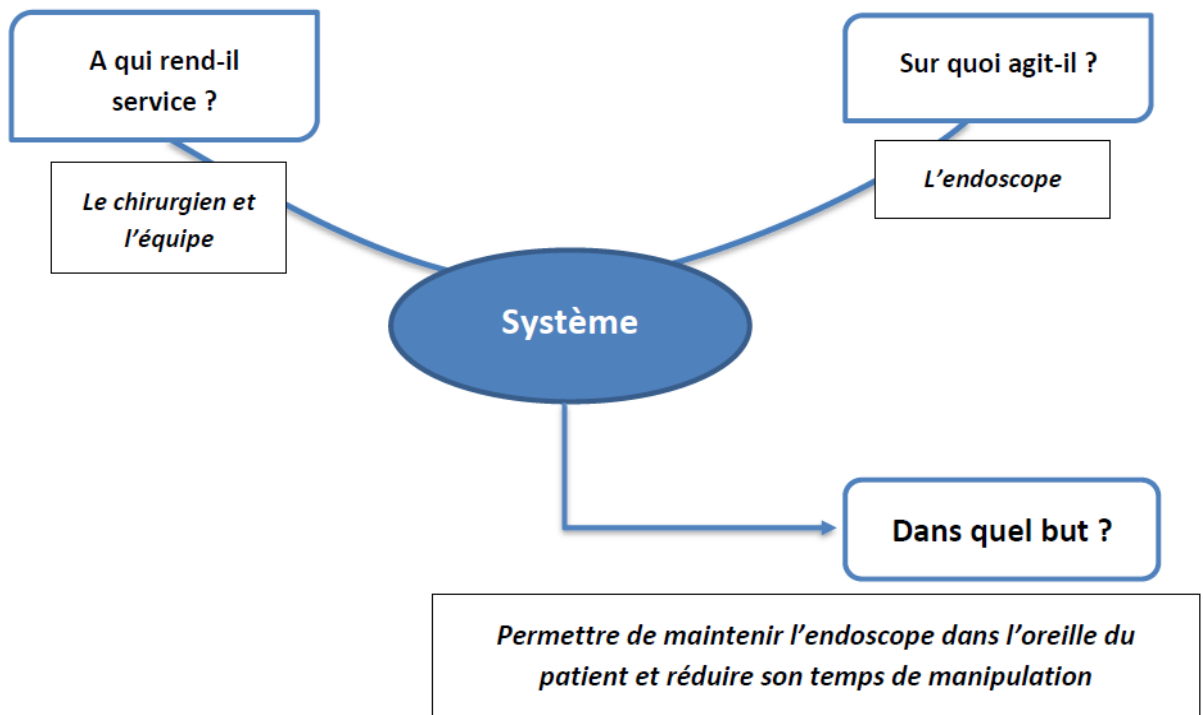

*Figure 3* : *Bête à cornes de l'analyse du système*

## 4   Analyse des différentes modalités de fixation

L'espace de travail étant défini, l'analyse fonctionnelle concerne l'environnement du patient dans la salle d'opération, et les contraintes de fixation du robot. Plusieurs modalités de fixation ont été envisagées.

### 1)   Au sol

Inspiré de l'état en l'art en laparoscopie, le robot peut être positionné sur un chariot roulant au sol. L'avantage est d'être mobile et aisé à transporter entre les salles de bloc. Ce type de fixation est répandu en salle d'intervention, et est déjà le cas pour le microscope par exemple. Cependant, cela majore l'encombrement, et réduit l'espace de travail autour du patient. Par ailleurs, cela nécessite une plus grande raideur entre le chariot et le sol, ce qui implique a priori un système plus lourd et volumineux.

### 2)   Au plafond

La base du robot serait déportée au plafond, ce qui permet de libérer l'espace au sol, comme pour les scialytiques par exemple. Cela permet une plus grande liberté de mouvement par rapport au patient, et libère de l'espace de travail. Cependant, cela contraint le système a n'être utilisé que dans une salle de bloc dédiée, ce qui peut être un problème dans certaines structures de soin. Par ailleurs, le placement est incertain au niveau des points d'attache au plafond par rapport à la table d'opération selon les salles de bloc, et cela rend plus difficile le dimensionnement et l'optimisation de l'espace de travail du robot.

### 3)   A la table d'opération

Inspiré du porte spéculum passif utilisé en otologie, ce système comporterait une fixation sur les rails placés de part et d'autre des tables d'opération. Cela permet un positionnement au plus près de l'espace de travail, et un dimensionnement du robot plus petit. Cependant, il faut être capable de limiter le volume du porte endoscope.

### 4)   Sur le patient

Il est également possible de venir fixer le robot porte-endoscope directement sur le patient. Cela a l'avantage de minimiser la taille du robot ; mais encombre nettement la place autour de l'espace de travail, déjà restreint, et probablement peut adapté à la chirurgie otologique.

### 5)   Sur le chirurgien

Certains dispositifs sont directement positionnés sur le chirurgien, comme les casques par exemple. Cela libère de l'espace et rend le positionnement plus intuitif. Cependant, le robot, de part son poids, peut devenir trop lourd et gênant lors d'interventions de durées souvent de 2 à 3 heures.

D'après les différentes possibilités, les choix les plus adaptés à la chirurgie otologique paraissent être la fixation sur rail de table d'opération, ou au sol. Cependant, il doit aussi être facilement transportable, donc sur une base roulante. Le choix final peut se faire en fonction de l'architecture du robot mais aussi en fonction de la complexité de la fixation ou de la nécessité de libérer de l'espace autour du patient pendant l'intervention.

## 5  Analyse des différentes modalités de contrôle

Pour réduire le temps de manipulation, il est nécessaire de connaître les modalités de contrôle du robot porte endoscope. Ce choix dépend de facteurs techniques, humains et éthiques. Une liste non exhaustive comprend les modalités suivantes.

### 6) Commande vocale :

Le chirurgien prononce des commandes préconçues, par exemple "Tourner 45 degrés X". Cela permet au chirurgien de ne pas utiliser de matériel supplémentaire et de conserver mains et pieds libres pour d'autres fonctions. Cependant, cela demande une phase d'apprentissage, et ces commandes peuvent entrer en conflit avec le dialogue entre le chirurgien et le personnel du bloc pendant l'opération. Cela nécessiterait donc l'utilisation d'une pédale d'activation, ce qui limite l'avantage a priori de cette méthode. Par ailleurs, il peut être difficile de préciser verbalement les amplitudes ou directions demandées au robot.

### 7) Manipulation par une tierce personne

Une tierce personne, c'est à dire autre que le chirurgien, manipule le porte-endoscope. Cependant, tout comme lors de la commande vocale, il peut être difficile, dans un espace restreint et anatomiquement complexe, de préciser à l'aide opératoire les directions voulues pour la manipulation du robot. Par ailleurs, cela nécessite une personne supplémentaire disponible durant toute la durée de l'opération, ce qui n'est pas possible dans toutes les structures.

### 8) Contrôle à l'aide d'une pédale ou d'un joystick

Si le chirurgien utilise un système de pédales pour guider les différentes mobilités du robot, ses mains restent libres pour effectuer l'opération, et en utilisant le pied gauche comme le pied droit, l'endoscope peut être orienté à sa guise tout au long de l'opération. Si le chirurgien utilise un joystick, il est obligé d'abandonner une main pour déplacer l'endoscope, ce qui reste réalisable s'il ne change pas trop souvent l'endoscope de configuration. Ces systèmes sont en général simples et intuitifs à mettre en place, mais cela diminue l'intérêt même du robot, qui vise à libérer les mains du chirurgien pendant l'opération. Au niveau des pieds, il existe déjà de multiples commandes qui limitent l'utilisation d'une pédale supplémentaire (aspiration, coagulation et moteur sont actionnés au pied).

### 9) Contrôle par mouvement de la tête

Une solution pourrait consister à équiper le chirurgien d'un casque comportant des accéléromètres. Le chirurgien peut alors contrôler le robot en déplaçant sa tête. Cependant, ce type de contrôle reste peu précis dans la gestion de la position et de la vitesse, et demande une longue phase d'apprentissage. Le développement est également couteux.

### 10) Utilisation d'un mode transparent

Le robot détecte quand un humain le manipule et passe alors dans un mode transparent, c'est à dire qu'il suit les mouvements qui lui sont imposés par l'humain et reste en position lorsqu'il perd le contact. Ce système est très simple et intuitif. Cependant, il nécessite la main du chirurgien ou de son assistant, et renvoit donc aux limites évoquées lors du contrôle par joystick.

### 11) Suivi d'un instrument chirurgical

Le robot utilise des algorithmes de vision pour détecter et tracker un outil. On peut penser à tracker l'aspiration qui est à priori toujours présente dans le champ de vision de l'endoscope et se déplace assez peu. Ce système serait intuitif et libèrerait les mains du chirurgien. Cependant, il faut pouvoir activer ou désactiver le tracking selon les situations.

Le choix de la solution de contrôle serait le suivi d'un instrument chirurgical, couplé au mode transparent pour le positionnement macroscopique du robot. Cependant, le développement d'un suivi optique peut être complexe à mettre en place et l'analyse de risque pourrait être difficile à réaliser si l'utilisateur n'est pas directement le pilote du porte endoscope. En cas d'accident, il sera difficile de remonter à l'origine de la défaillance.

# 6 Conclusion

L'analyse du besoin réalisée spécifiquement pour la chirurgie otologique a permis de définir l'espace de travail. Cet espace de travail est de volume restreint, mais peut être agrandi par un fraisage des structures osseuses. Le choix de la solution de fixation pourrait être au sol ou à la table, et a un impact sur le choix futur de l'architecture. Le choix de la solution de contrôle est clairement en faveur d'un suivi d'instrument grâce à une analyse de l'image en temps réel, afin de libérer les mains du chirurgien. Cette analyse fonctionnelle prérequis pour la conception d'un robot assistant en chirurgie otologique, permet de poser les premières exigences liées à cette chirurgie auxquelles le robot devra répondre.

# 7 Références